\documentclass{article}

\usepackage[nonatbib, final]{neurips_2022}

\usepackage{subfig}
\usepackage{microtype}
\usepackage{graphicx}
\usepackage{wrapfig}
\usepackage{booktabs} 

\usepackage{xr}
\makeatletter
\newcommand*{\addFileDependency}[1]{
  \typeout{(#1)}
  \@addtofilelist{#1}
  \IfFileExists{#1}{}{\typeout{No file #1.}}
}
\makeatother

\newcommand*{\myexternaldocument}[1]{%
    \externaldocument{#1}%
    \addFileDependency{#1.tex}%
    \addFileDependency{#1.aux}%
}

\myexternaldocument{appendix}

\usepackage[
backend=bibtex,
url=false,isbn=false,doi=false,
date=year,
hyperref=auto,
style=alphabetic,
natbib=true,
sorting=nty,
firstinits=true,
maxnames=2, maxbibnames=10,
]{biblatex}

\bibliography{./reference.bib}

\usepackage{appendix}
\usepackage[figuresright]{rotating}
\usepackage{amsmath,amssymb,amscd,amsthm}
\usepackage{dcolumn}
\usepackage{bm}
\usepackage{ifthen}
\usepackage{rays_defs_18}
\usepackage{commath}
\usepackage{bbm}
\usepackage{wrapfig}
\usepackage{algorithm,algorithmic} 

\usepackage{xcolor}
\usepackage{xspace}
\usepackage{mathtools}
\usepackage{paralist}
\usepackage{array}

\newcommand{\ignore}[1]{}

\usepackage{hyperref}

\newcommand\method{{\em ScatterSample}\xspace}
\newcommand\samplmethod{DiverseUncertainty\xspace}

\newcommand{\theory}{Theoretical Guarantees\xspace}

\newcommand{\insight}{Insight\xspace}
\newcommand{\effectiveness}{Effectiveness\xspace}

\usepackage{hyperref}

\usepackage{amsmath}
\usepackage{amssymb}
\usepackage{mathtools}
\usepackage{amsthm}

\usepackage[capitalize,noabbrev]{cleveref}

\theoremstyle{plain}
\newtheorem{theorem}{Theorem}[section]

\theoremstyle{definition}

\theoremstyle{remark}

\usepackage[textsize=tiny]{todonotes}

\title{\method: Diversified Label Sampling for Data Efficient Graph Neural Network Learning}

\author{%
Zhenwei Dai  \\
Rice University \\
Houston, Texas, USA \\
\And
Vasileios Ioannidis \\
Amazon Web Services \\
Santa Clara, California, USA \\
\AND
Soji Adeshina \\
Amazon Web Services \\
Santa Clara, California, USA \\
\And
Zak Jost \\
Amazon Web Services \\
Santa Clara, California, USA \\
\And
Christos Faloutsos \\
Amazon Web Services \\
Santa Clara, California, USA \\
\And 
George Karypis \\
Amazon Web Services \\
Santa Clara, California, USA \\
}

\begin{document}

\maketitle

\begin{abstract}
What target labels are most effective for graph neural network (GNN) training? In some applications where GNNs excel-like drug design or fraud detection, labeling new instances is expensive. We develop a data-efficient active sampling framework, ScatterSample, to train GNNs under an active
learning setting. ScatterSample employs a sampling module termed DiverseUncertainty to collect instances with large uncertainty from different regions of the sample space for labeling. To ensure diversification of the selected nodes, DiverseUncertainty clusters the high uncertainty nodes and selects the representative nodes from each cluster. Our ScatterSample algorithm is further supported by rigorous theoretical analysis demonstrating its advantage compared to standard active sampling methods that aim to simply maximize the uncertainty and not diversify the samples. In particular, we show that ScatterSample is able to efficiently reduce the model uncertainty over the whole sample space. Our experiments on five datasets show that ScatterSample significantly outperforms the other GNN
active learning baselines, specifically it reduces the sampling cost by up to \bm{$50\%$} while achieving the same test accuracy. 
\end{abstract}

\section{Introduction}

How to spot the most effective labeled nodes for GNN training?
Graph neural networks (GNN)~\cite{kipf2016semi,velivckovic2017graph,wu2019simplifying} 
which employ non-linear and parameterized feature 
propagation~\cite{zhu2002learning} to compute graph representations, 
have been widely employed in a broad range of learning tasks and achieved state-of-art-performance in node classification, link prediction and graph classification. 
Training GNNs for node classification in the supervised learning setup typically requires a large number of labeled examples such that the GNN can learn from diverse node features and node connectivity patterns. However, labeling costs can be expensive which inhibits the possibility of acquiring a large number of node labels.  For example, the GNNs can be used to assist the drug design. However, evaluating the properties of a molecule is time consuming. It usually takes one to two weeks for evaluation using the current simulation tools, not to mention the cost spent on the laboratory experiments.

Active learning (AL) aims at maximizing the generalization performance under a constrained labeling budget~\cite{settles2009active}. AL algorithms choose which training instances to use as labeled targets to maximize the performance of the learned model. Previous research in AL algorithms for GNN training can be categorized with respect to whether the AL methods take into account the model weights (model aware) or can be applied to any model (model agnostic). Model agnostic algorithms label a representative subset of the nodes such that the labeled nodes can cover the whole sample space~\cite{wu2019active, zhang2021grain}. Model aware AL algorithms leverage the GNN model to compute the node uncertainty, which combines both the input features and graph structure~\cite{cai2017active, gao2018active}. 
AL then picks the nodes with the highest uncertainty.

However, maximizing the uncertainty of the labeled nodes may not balance the exploration and exploitation of the classification boundary~\cite{kirsch2019batchbald}. For example, if there exist a group of nodes close to the classification boundary but are clustered in a small region of the graph, just labeling the most uncertain nodes could only explore that specific region of the classification boundary, while others are ignored, and the classification boundary is not well explored.
Thus, our first main contribution
is to simultaneously consider the node uncertainty and the diversification of the uncertain nodes over the sample space.  

\paragraph{Challenges of diversifying uncertain nodes}
\label{para;challenges}
Graph data present additional challenges to diversify the uncertain nodes. Diversification requires modeling the sample space using carefully selected representations for the nodes. However, there are two challenges of a suitable node representations.
\begin{itemize}
\item[Challenge 1:] Sample space for graph data requires a representation which takes both the graph structure and node features into account (see section sec~\ref{sec:diversify_uncertain}).
\item[Challenge 2:] The representation should be robust to the model trained so far, and not be biased by the limited amount of available labels.
\end{itemize}

\paragraph{Our approach}
We develop {\method} for data-efficient GNN learning. 
\method allows us to explore the classification boundary while exploiting the nodes with the highest uncertainty. To diversify the uncertain samples on graph-structured data,  \method includes a {\em \samplmethod} module to address the two challenges above, which clusters the uncertain nodes representations over the whole sample space.

\begin{wrapfigure}{r}{0.4\textwidth}
\centering
{\includegraphics[width=0.4\textwidth]{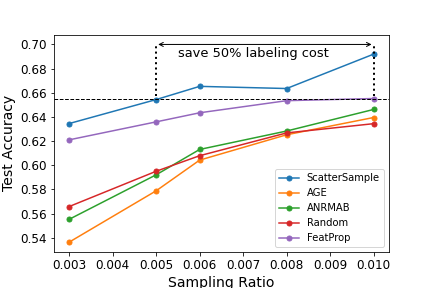}} 
\caption{\textbf{\method wins:}   
test accuracy 
vs. sampling ratio
on the ogbn-products dataset (62M edges). 
\label{fig:acc_compare_ogbn}}
\end{wrapfigure}

\paragraph{Our Contributions} 
The contributions of our work are the following.
\begin{compactitem}
    \item {\bf \insight:} 
    \method is the first method that proposes and implements diversification of the uncertain samples for data efficient GNN learning. 
    \item {\bf \effectiveness:} We evaluate \method on five different graph datasets, where \method  saves up to 
     \bm{$50\%$} labeling cost,
    while still achieving the same test accuracy with
    state-of-the-art baselines.
    \item {\bf \theory}: Our theoretical analysis proves the superiority of \method over the standard, uncertainty-sampling method (see Theorem~\ref{theorem_1}). Simulation results further confirm our theory. 

\end{compactitem}

\section{Related Work}

This section will review the uncertainty based active learning research and implementation of active learning in GNNs.

\paragraph{Active Learning (AL):}
Active learning aims at selecting a subset of training data as labeling targets such that the model performance is optimized~\cite{settles2009active, hanneke2014theory}. Uncertainty sampling is one major approach of active learning, which labels a group of samples to maximally reduce the model uncertainty. To achieve this goal, uncertainty sampling selects samples around the decision boundary~\cite{tur2005combining}. Uncertainty sampling has also been applied to the deep learning field, and researchers have proposed different methods to measure the uncertainty of samples. For example, \citet{ducoffe2018adversarial} developed a margin based method which uses the distance from a sample to its smallest adversarial sample to approximate the distance to the decision boundary. 

\paragraph{AL and GNNs:}
AL with GNNs requires to consider the graph structure information into the node selection. \citet{wu2019active} uses the propagated features followed by K-Medoids clustering of nodes to select a group of representative instances. \citet{zhang2021grain} measures importance of nodes through combining the diversity and influence scores. However the above approaches do not account for the learned GNN model, which may limit the generalization performance. 
Uncertainty sampling has also been implemented to select nodes. \citet{cai2017active} propose to use a weighted average of the node uncertainty, graph centrality and information density scores. \citet{gao2018active} further propose a different approach to combine the three features with multi-armed bandit techniques. Although useful, these approaches aim choose nodes with the highest uncertainty and may be challenged if the selected nodes are clustered in a small region of the graph, which will not provide good graph coverage. Our work addresses this limitation by diversifying the selected nodes based on the graph structure.



\section{Preliminaries}
\paragraph{Problem Statement } Given a graph $\setG = (\setV, \setE)$, where $\setV$ is the set of nodes  with $N= |\setV|$ nodes and $\setE$ is the set of edges. The set of nodes is divided into the training set $\setV_{train}$, validation set $\setV_{valid}$ and testing set $\setV_{test}$. Each node $v_n \in \setV$ is associated with a feature vector $\vx_n \in \reals^d$ and a label $y_n \in \{1,2, \ldots, C\}$. Let $\mX \in \reals^{N \times d}$ be the feature matrix of all the nodes in the graph, where the $i$-th row of $\mX$ corresponds to $v_n$, $\vy = (y_1, y_2, \ldots, y_n) \in \reals^n$ is the vector containing all the labels. To learn the labels of the nodes, we train a GNN model $M$ which maps the graph $\setG$ and $\mX$ to the the prediction of labels $\hat{\vy}$. 
\paragraph{Active Learning:} Active learning picks a subset of nodes $S \subset \setV_{train}$ from the training set and query their labels $\vy_S$. A GNN model $M_S$ is trained with respect to the feature matrix $\mX$ and $\vy_S$. Given the sampling budget $B$, the goal of active learning is to find a set $S$ ($|S| \leq B$) such that the generalization loss is minimized, i.e.
\[
\argmin_{S:|S| \leq b} \EX[v_n \in \setV_{test}]{\ell(y_n, f(\vx_n|\setG, M_S))}.
\]
\subsection{Graph neural networks and message passing}
In this section we present the basic operation of the GNN at layer $l$. With the message passing paradigm, the GNN layer updates for most GNN models can be interpreted as message vectors that are exchanged among neighbors over the edges and nodes in the graph.

For the following let $\mathbf{h}^{(l)}_v\in\mathbb{R}^{d_1}$ be the hidden representation for node $v$ and layer $l$. Consider $\phi$ that is a message function combining the hidden representations for nodes $v,u$. Next, using the message vectors for neighboring edges the node representations are updated as follows 
\begin{align} \mathbf{h}_v^{(l+1)} = \psi \left(\mathbf{h}_v^{(l)}, \rho(\lbrace \phi ( \mathbf{h}_v^{(l)}, \mathbf{h}_u^{(l)}) : ({u}, {v}) \in  \setE \rbrace ) \right) 
\end{align} 
where $\rho$ is a reduce function used to aggregate the messages coming from the neighbors of $v$ and $\psi$ is an update function defined on each node to update the hidden node representation for layer $l+1$.  By defining $\phi, \rho, \psi$ different GNN models can be instantiated~\cite{kipf2016semi,defferrard2016convolutional,bronstein2017geometric,ioanniditensorgcnstspjournal}. These functions are also parameterized by learnable matrices that are updated during training.

\section{Proposed method: \method}

We propose the \method algorithm, which dynamically samples a set of diverse nodes with large uncertainties in order to more efficiently explore the classification boundary during GNN training. At each round, our method calculates the uncertainty for all nodes with the GNN model trained so far. Then, \method clusters the top uncertain nodes and selecting nodes from each cluster to obtain diverse samples. The labels of the selected nodes are queried and used as supervision to continue training the GNN model for the next round. This section explains our method in detail.

\subsection{Selecting the uncertain nodes}

The uncertainty of a node is measured by the information entropy. Given a trained GNN model at the $t$-th sampling round, \method first computes the information entropy $\phi_{entropy}(v_n)$ of nodes in $\setV_{train}$ based on the current GNN model, i.e.
\begin{align}
    \phi_{entropy}(v_n) = -\sum^{C}_{j=1} \log(\PR{Y_{n}=j \mid \setG, \mX, M})  \PR{Y_{n}=j \mid \setG, \mX, M}    
    \label{eq:entropy}
\end{align}
where $\PR{Y_{n}=j \mid \setG, \mX, M}$ is probability that node $v_n$ belongs to class $j$ given the GNN model $M$. Then, \method ranks all the nodes in order of decreasing uncertainty, and picks the ones with the largest information entropy into a candidate set $\C_t\subset\setV_{train}$. Different than traditional AL techniques that select training targets solely based on uncertainty, we then move on to pick a diverse subset of the uncertain nodes over the sampling space.

\subsection{Diversifying uncertain nodes}
\label{sec:diversify_uncertain}
Our goal is to ensure the diversity of selected nodes for labeling, by exploring the node distribution over the sample space. At this point naturally, the question arises \emph{How to model the sample space?} We need a representation for nodes to define the space, based on which we could measure the samples' distances. A straightforward approach is to use the GNN embedding space since the classification boundary is directly depicted there. However, GNN embeddings fail to address the two challenges in the introduction section.   

First, with active learning, a limited number of labeled nodes are available in the initial stages. Hence, only the already labeled nodes may have reliable GNN embeddings and biased subsequent samples. Second, GNN embeddings for node classification may not carry enough information for diversification. GNNs usually do not have an MLP layer connecting to the output. The final GNN outputs of uncertain nodes are not diverse enough since the high uncertain nodes may have similar class probabilities (class probabilities close to uniform). Conversely, embeddings of intermediate GNN layers may have an appropriate dimension but lack information of the expanded ego-network.

These drawbacks are confirmed in Sec.~\ref{sec:ablation}, where we show that using GNN embeddings as proxy representations leads to a performance drop. Moreover, different from other machine learning problems, the nodes are correlated with each other, and we also need to take the graph structure into account when diversifying the samples. Hence, to address all these considerations we will employ a $k$-step propagation of the original node features based on the graph structure as a proxy representation for the nodes. The $k$-step propagation of nodes $\mX^{(k)} = (\vx^{(k)}_1, \vx^{(k)}_2, \ldots, \vx^{(k)}_N)$ is defined as follows
\begin{equation}
\mX^{(k)} := \mS\mX^{(k-1)}
	\label{eq:sem}
\end{equation}
where $\mS$ is the normalized adjacency matrix, and $\mX^{(0)}$ are the initial node features. The operation in equation~\ref{eq:sem} is efficient and amenable to a mini-batch implementation. Such representations are well-known to succinctly encode the node feature distribution and graph structure. Next, we calculate the proxy representations for the candidate high uncertainty nodes in the set $\mathcal{C}_t$. To maximize the diversity of the samples, we cluster the proxy representations in $\mathcal{C}_t$ using $k$-means++ into $B_t$ clusters~\cite{arthur2006k}, and select the nodes closest to the cluster centers for labeling, by using the $L_2$ distance metric. One node from each cluster is selected that amounts to $B_t$ samples.

\begin{algorithm}[ht]
\caption{\method Algorithm}
\label{alg:algo1}
\begin{algorithmic}[1]
  \STATE \textbf{Input}: $\setV_{train}$, GNN model $M$, number of propagation layers $k$, number of sampling round $T$,  sampling redundancy $r$, initial sampling budget $B_0$ and total sampling budget $B$.
  \STATE Initialize  $S = \emptyset$
  \STATE Compute $\mathbf{x}^{(k)}_{n}~\forall n \in \setV_{train}$ as in equation~\ref{eq:sem}.
  \newline
  \STATE \textbf{Initial Sampling:} 
  \STATE Use $k$-means++ to cluster $\{\mathbf{x}^{(k)}_{n}\}$ into $B_0$ clusters. 
  \STATE Add a node closest to the cluster center per cluster to $S$.
  \STATE Query the labels of nodes $v_n \in S$, denoted by $\vy_S$.
  \STATE Train model $M$ using $(\vy_S, \mX, \setG)$. 
  \newline
  \STATE \textbf{Dynamic Sampling:}
  \STATE Initialize sampling round $t=1$
  \WHILE{$t < T$}
  \STATE Let $B_t = \min(B-|S|, (B-B_0)/T)$
  \STATE Use the \textbf{\samplmethod} algorithm to select $S_t$ 
  \STATE Query the labels of $S_t$, and update $S = S \cup S_t$.
  \STATE Train model $M$ over $(\vy_S, \mX, \setG)$. Update $t = t + 1$.
  \ENDWHILE
\end{algorithmic}
\end{algorithm}

Clearly, the size of the candidate set $|\C_t| \geq B_t$, however deciding how many candidate nodes to choose from is important. We parameterize the size as a multiple of the selected nodes namely $|\C_t|=rB_t$, where $r>1$ is the sampling redundancy. If $r$ is too small, the selected nodes are closer to the classification boundary (have larger information entropy) but the nodes selected may not be diverse enough. On the other hand, if $r$ is too large, the set will be diverse, but the selected nodes may be far away from the classification boundary. Therefore, it is critical to pick a suitable $r$ to achieve a sweet point between diversity and uncertainty. We leave the discussion of choosing $r$ to Sec.~\ref{sec:ablation}. Besides empirical validation with experiments in five real datasets (see Sec.~\ref{sec:exp}), our diversification approach is theoretically motivated (see Sec.~\ref{sec:theory}).

\begin{algorithm}[ht]
\caption{\samplmethod Algorithm}
\label{alg:maxdiversity}
\begin{algorithmic}[1]
  \STATE \textbf{Input}: $\setV_{train}$, $\{\mathbf{x}^{(k)}_{n}~~\forall n\in \mathcal{C}_t\}$, $r$, $B_t$
  \STATE Compute $\phi_{entropy}(v)~~\forall v \in \setV_{train}$; see equation~\ref{eq:entropy}).
  \STATE $\mathcal{C}_t$ $\leftarrow$ \{$rB_t$ nodes with largest $\phi_{entropy}(v)$\}.
  \STATE Use $k$-means++ to cluster the $\vx^{(k)}_n$ (for all $ n\in\mathcal{C}_t$) into $B_t$ clusters.
  \STATE $S_t \leftarrow \emptyset$
  \FOR{$j=1,2,\ldots,B_t$}
  \STATE Compute the cluster center $\vv_j$ of cluster $j$
  \STATE Pick node $x \leftarrow \argmin_{n \in \mathcal{C}_t} \norm[2]{\vx^{(k)}_n - \vv_j}$
  \STATE $S_t \leftarrow S_t \cup \{x\}$
  \ENDFOR
  \STATE Return $S_t$
\end{algorithmic}
\end{algorithm}

The pseudo code of \method is shown in Algorithm~\ref{alg:algo1}. \method is a multiple rounds sampling scheme, which includes an initial sampling step and dynamic sampling steps. \method first computes the $k$-step features propagation of all the nodes in the training set using equation~\ref{eq:sem}, and clusters them into $B_0$ clusters, where $B_0$ is the initial sampling budget. Then, \method picks the nodes closest to the cluster centers as the initial training samples and queries their labels. The purpose of clustering $k$-step feature propagations is to enforce the initial training set to spread out over the whole sample space. It is also helpful to explore the classification boundary since if the initial sampled nodes are not diverse enough, we cannot picture the classification boundary of the regions that are far away from the initial training samples. \method repeats the dynamic sampling described in Algorithm~\ref{alg:maxdiversity} until the sampling budget $B$ is exhausted. The next section fortifies our diversification method with theoretical guarantees.

\section{Theoretical analysis}
\label{sec:theory}

In Sec.~\ref{sec:ablation}, we have shown that \samplmethod is significantly better than Uncertainty algorithm. In this section, we provide theoretical analysis and simulation results to demonstrate the benefits of \samplmethod and explains why MaxUncertainty algorithm may fail.
The results presented here give a theoretical basis for the superiority of our method as established in the experiments in Section~\ref{sec:exp}.

\subsection{Analysis setup}
\label{sec:assump}
For the analysis, we employ the  Gaussian Process (GP) model~\cite{o1978curve}. GP models offer a flexible approach to model complex functions and are robust to small sample sizes~\cite{seeger2004gaussian}. Moreover, the uncertainty of the prediction can be easily computed using a GP model. Neural network models and GNNs interpolate the observed samples, while GPs provide a robust framework to interpolate samples, that is amenable to analysis.

Assume the label $y_i \in \reals$ is dependent on the propagated features $\vx^{(k)}_i$ through a GP model.  
The label $y_i$ is modeled by a Gaussian Process, where $(\vy \mid \mX^{(k)}) \sim N(\bm{1}\mu, \mK(\mX^{(k)}))$ and $\mK(\mX^{(k)})$ is the Gaussian kernel matrix. The kernel is parameterized by $\mK_{ij}(\mX^{(k)}) = K(\vx^{(k)}_i, \vx^{(k)}_j) = \exp\left(-\frac{1}{2}(\vx^{(k)}_i - \vx^{(k)}_j)^T\Sigma^{-1}(\vx^{(k)}_i - \vx^{(k)}_j)\right)$, where $\Sigma = diag(\theta_1, \theta_2, \ldots, \theta_d)$. 
Consider that the sample space of $\vx^{(k)}$ can be clustered into $m$ clusters $\mc S_1, \mc S_2, \ldots, \mc S_m$, and denote the cluster centers as $\vc_1, \vc_2, \ldots, \vc_m$. Without loss of generality, denote the radius of the cluster, $d_1 \leq d_2 \leq d_3 \leq \cdots < d_m$. The clusters are well separated and the distance between the cluster centers are larger than $\delta$, i.e. $\min_{i \neq j} \norm[2]{\vc_i - \vc_j}_2 \geq \delta$ ($\delta > 2d_m$).  Moreover, we consider that there does not exist a cluster dominating the sample space, $d^2_m \leq \tau \sum_{j=1}^{m-1} d^2_j$ and the samples are uniformly distributed over the clusters. 

\subsection{MaxUncertainty vs \samplmethod}

Here, we show that \samplmethod could significantly achieves smaller mean squared error (MSE) compared to MaxUncertainty.  Without loss of generality we consider $m$ clusters and the following definitions.
\begin{itemize}
    \item \textbf{MaxUncertainty} Select $2m$ most uncertain samples.
    \item \textbf{\samplmethod} Select the $2$ most uncertain samples from each cluster.
\end{itemize}

\begin{figure}[ht]
\centering
\subfloat[MaxUncertainty]{\includegraphics[width=0.3\textwidth]{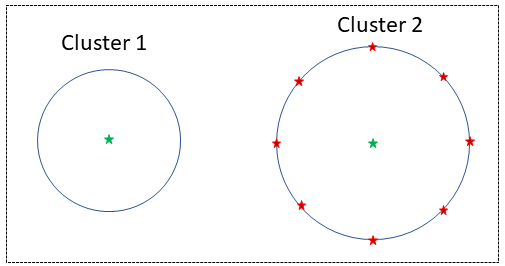}} 
\subfloat[\samplmethod]{\includegraphics[width=0.3\textwidth]{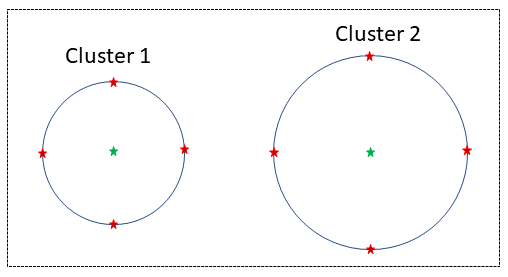}}
\caption{The area enclosed by the blue circles is the sample space of propagated features (2D case). The green stars are sampled nodes during initial sampling (cluster center). The red stars are the sampled nodes during uncertainty sampling. (a) MaxUncertainty picks the nodes with largest uncertainty, which is equivalent to sampling the boundary of cluster 2. (b) \samplmethod diversifies the clustered nodes, and samples the boundary of both clusters.}
\label{fig:diversity_uncertainty}
\end{figure}

Before presenting the theory we illustrate the operation of our method and \textbf{MaxUncertainty} in~Figure~\ref{fig:diversity_uncertainty}. \method first clusters the samples on the propagated feature space (blue circles in Figure~\ref{fig:diversity_uncertainty}), and selects the nodes closest to the cluster centers for initial training (green stars in Figure~\ref{fig:diversity_uncertainty}).  Then, during the dynamic sampling steps, we compute the uncertainty using equation~\ref{GP:infer}. MaxUncertainty approach will select the nodes with the largest uncertainty. Under our setup, it is equivalent to sample nodes at the boundary of the largest cluster since the distance to the cluster center is the most important factor of uncertainty (Figure~\ref{fig:diversity_uncertainty}(a)). While \samplmethod will diversify the high uncertainty nodes, which is equivalent to sample from the boundary of each cluster (Figure~\ref{fig:diversity_uncertainty}(b)). The red stars of Figure~\ref{fig:diversity_uncertainty} show the nodes labeled during the uncertainty sampling stage. Since MaxUncertainty algorithm only labels the nodes in cluster 2, cluster 1 is ignored the prediction uncertainty of cluster 2 cannot be reduced. On the contrary, \samplmethod samples nodes from both cluster 1 and 2. Thus, it could reduce the prediction uncertainty in both clusters. 

Then, the following theorem quantifies the relationship of the MSEs of both algorithms under the setup of Sec.~\ref{sec:assump}.  



\begin{theorem}
\label{theorem_1}
Consider a case where feature dimension $d=1$. With the above notation and assumptions, let $r_i = \Exp{-\frac{d^2_i}{2\theta}}$. If we satisfy $d^2_m \geq d^2_{m-1} + 4\log\theta$ and $\delta \geq d_m + \max\left(\sqrt{d^2_m + \theta \log(9m)}, 2\theta\log(\frac{3\sqrt{m}}{1-r_m})\right)$, we have
\[
\frac{MSE(f(x)|\text{MaxUncertainty})}{MSE(f(x)|\text{\samplmethod})}
\geq \frac{1}{2(1+\tau)} \frac{1+r^2_m}{1-r_m} - \frac{8}{3}
=  \frac{1}{\tau+1}O(\theta).
\]
\end{theorem}


Theorem~\ref{theorem_1} suggests that when the GP function is smooth enough (large $\theta$), the MaxUncertainty will have larger MSE than the MaxDiversity algorithm. A large $\theta$ suggests a close correlation between the labels of the nodes that are close to each other. It is also common for most of the graph datasets where samples clustered together usually have similar labels. Thus, \samplmethod can achieve a smaller MSE in this case. 

\section{Experiments}
\label{sec:exp}

We evaluate the performance of \method  on five different datasets.

\paragraph{Datasets}
We evaluated the different methods on the Cora, Citeseer, Pubmed, Corafull~\cite{kipf2016semi}, and ogbn-products~\cite{hu2020ogb} datasets (Table~\ref{tab:data}). Besides the ogbn-products, we do not keep original data split of the training and testing set. For the nodes that are not in the validation or testing sets (the validation and testing sets follows the split in the dgl package ``dgl.data''~\cite{wang2019dgl}), we will add them to the training set. The labels can only be queried from the training set. 

\begin{table}[H]
\centering
\caption{Statistics of graph datasets used in experiments.} \label{tab:data}
\renewcommand{\arraystretch}{0.8}
\begin{tabular}{ccccc}
\toprule
Data          & \# Nodes  & \# Train Nods & \# Edges   & \# Classes \\ \midrule
Cora          & 2,708     & 1,208         & 5,429      & 7          \\ 
Citeseer      & 3,327     & 1,827         & 4,732      & 6          \\ 
Pubmed        & 19,717    & 18,217        & 44,328     & 3          \\ 
Corafull      & 19,793    & 18,293        & 126,842    & 70         \\ 
ogbn-products & 2,449,029 & 196,615       & 61,859,149 & 47         \\ \bottomrule
\end{tabular}
\end{table}

\paragraph{Baselines}
For different sampling budget $B$, we compare the test accuracy of \method with the following graph active learning baselines:
\begin{compactitem}
\item \emph{Random sampling}. Select $B$ nodes uniformly at random from $\setV_{train}$. 
\item \emph{AGE}~\cite{cai2017active}: AGE computes a score which combines the node centrality, information density, and uncertainty, to select $B$ nodes with the highest scores.
\item \emph{ANRMAB}~\cite{gao2018active}: ANRMAB learns the combination weights of the three metrics used by AGE with multi-armed bandit method. 
\item \emph{FeatProp}: FeatProp~\cite{wu2019active} clusters the feature propogations into $B$ clusters and pick the nodes closest to the cluster centers.
\item \emph{Grain}:~\cite{zhang2021grain} score the node by the weighted average of the influence score and diversity score. And select the top $B$ nodes with largest node scores. Grain includes two different approaches of selecting nodes, Grain (ball-D) and Grain (NN-D).
\item \emph{\method}: For the sample scale graph dataset (Cora, Citeseer), we set the initial sampling budget to $3\% \cdot |\setV_{train}|$ and sample $1\% \cdot |\setV_{train}|$ each round during the dynamic sampling period. For medium scale datasets (Pubmed and Corafull), we set the initial sampling budget to $1\% \cdot |\setV_{train}|$ and sample $0.5\% \cdot |\setV_{train}|$ each dynamic sampling round. For the large scale dataset (ogbn-products), initial sampling budget is $0.2\% \cdot |\setV_{train}|$, and each dynamic sampling round selects $0.05\% \cdot |\setV_{train}|$ nodes.
\end{compactitem}

\paragraph{GNN setup}
We train a $2$-layer GCN network with hidden layer dimension = $64$ for Cora, Citeseer and Pubmed, and $=128$ for Corafull and obgn-products. To train the GNN, we follow the standard random neighbor sampling where for each node~\cite{hamilton2017inductive}, we randomly sample 5 neighbors for the convolution operation in each layer. We use the function in ``dgl'' package to train the GNNs~\cite{wang2019dgl}.

\subsection{Performance Results}

We compare the performance of different active graph neural network learning algorithms under different labeling budgets ($B$). We parameterize the labeling budget $B$ equal to a certain proportion of the nodes in the training set ($B = r|\setV_{train}|$).
For Cora and Citeseer, we vary $r$ from $5\%$ to $15$\% in increment of $2\%$; for Pubmed and Corafull, $r$ is varied from $3\%$ to $10\%$; for ogbn-product dataset, we vary the $r$ from $0.3\%$ to $1\%$. The performance of the active learning algorithms are measured with the test accuracy.

\begin{figure}[H]
\centering
\subfloat[Cora]{\includegraphics[width=0.33\textwidth]{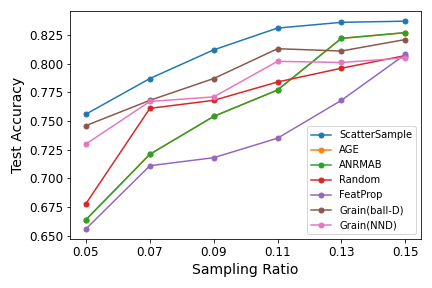}}
\subfloat[Citeseer]{\includegraphics[width=0.33\textwidth]{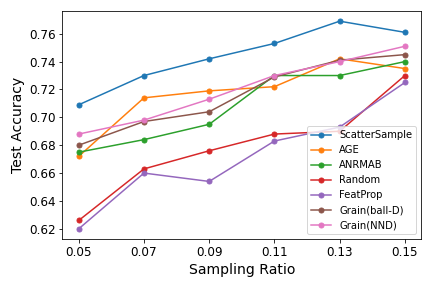}} \hfill
\subfloat[Pubmed]{\includegraphics[width=0.33\textwidth]{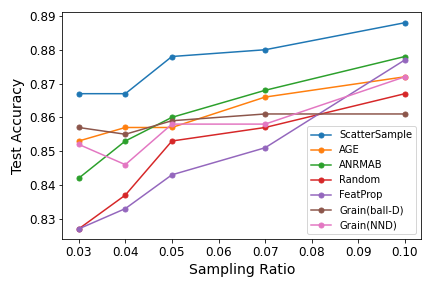}} 
\subfloat[Corafull]{\includegraphics[width=0.33\textwidth]{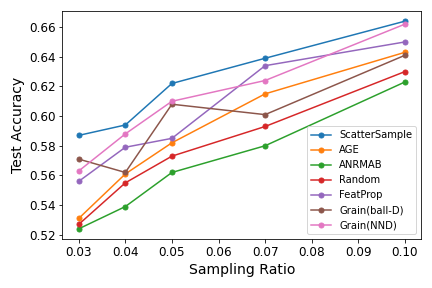}}
\caption{{\bf \method (blue), wins consistently:}  Comparison of the test accuracy of active GNN learning algorithms at different labeling budget. The $x$-axis shows \# labeled nodes/\# nodes in training set.}
\label{fig:acc_compare}
\end{figure}

\paragraph{Accuracy}
Figure~\ref{fig:acc_compare} shows the test accuracy of baselines trained on different proportions of the selected nodes. \method improves the test accuracy and consistently outperforms other baselines in all the datasets. In Citeseer, \method requires $9\%$ of the node labels to achieve test accuracy $74.2\%$, while the best alternative baselines ``Grain (ball-D)'' and ``Grain (NN-D)'' need to label $15\%$ of nodes to achieve similar accuracy, which corresponds to a $40\%$ savings of the labeling cost. Similarly, in PubMed and ogbn-products, \method achieves a $50\%$ labeling cost reduction compared to the best alternative baseline.

\paragraph{Efficiency}
Here, we compare the computation time among the methods that use the graph structure and node features to select the samples namely, \method, ``Grain (ball-D)'' and ``Grain (NN-D)''. We use the ogbn-products dataset to perform comparisons. \method takes less than $8$ hours to determine the labeling nodes and train the GNN, while the Grain algorithm requires more than $240$ hours. Grain requires $\mathcal{O}(n^2)$ complexity to calculate the scores of all nodes, which is prohibitive complexity in large graphs.

\paragraph{Complexity analysis}
The computation complexity of \samplmethod is $O(|E| + r*B^2_t)$. It is because \method includes two parts: 1) computing the node representations with complexity $O(|E|)$ where $|E|$ is the number of edges and 2) cluster the the uncertain nodes where the complexity is $O(rB^2_t)$. Since both $r$ and $B_t$ are small, $rB^2_t < |E|$, our method does not add a lot of extra burden compared to the model training time.

\subsection{Ablation Study}
\label{sec:ablation}

The MaxDiversity algorithm of \method needs to determine the size of candidate set $\mathcal{C}_t$ before selecting a subset $S_t$ from $\mathcal{C}_t$ for labeling. Hence, sampling redundancy $r$ and the clustering algorithm to cluster the nodes in $\mathcal{C}_t$ will affect the performance of \method. In this section, we will evaluate the effect of both factors. 

\begin{figure}[ht]
\centering
\subfloat[Cora]{\includegraphics[width=0.33\textwidth]{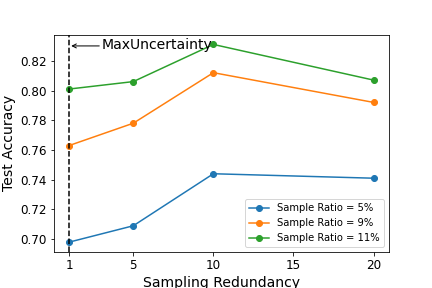}} 
\subfloat[Citeseer]{\includegraphics[width=0.33\textwidth]{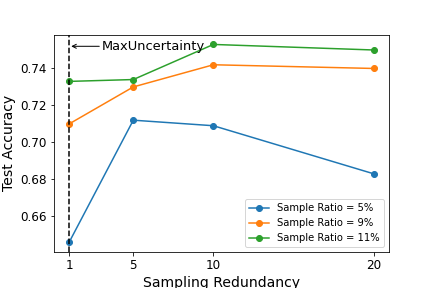}} 
\subfloat[Pubmed]{\includegraphics[width=0.33\textwidth]{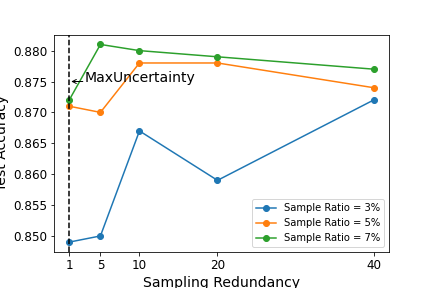}} 
\caption{Compare the performance under different sampling redundancy $r$. When $r=1$, \samplmethod reduces to MaxUncertainty method.}
\label{fig:sample_redundancy}
\end{figure}

\paragraph{Sampling redundancy $r$:} Recall from algorithm~\ref{alg:algo1}, the sampling redundancy $r$ controls the relative size of candidate set $\mathcal{C}_t$ to size of sampled node $S_t$. When $r=1$, \method reduces to the standard MaxUncertainty algorithm. And figure~\ref{fig:sample_redundancy} shows that the sampling the most uncertain nodes is significantly worse than \samplmethod. For the Citeseer dataset, \samplmethod can outperform MaxUncertainty by over $7\%$ when sampling ratio is $5\%$.
Therefore, to achieve a good test accuracy, $r$ should be carefully selected. Figure~\ref{fig:sample_redundancy} suggests that as $r$ increases, the test accuracy quickly boosts at the early stage, and then decreases slowly.

\begin{figure}[ht]
\centering
{\includegraphics[width=0.35\textwidth]{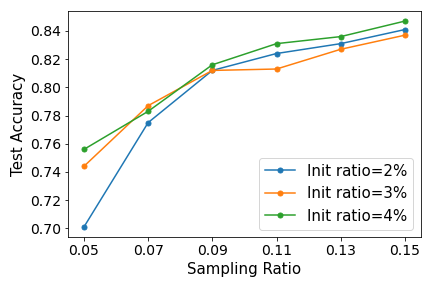}} 
{\includegraphics[width=0.35\textwidth]{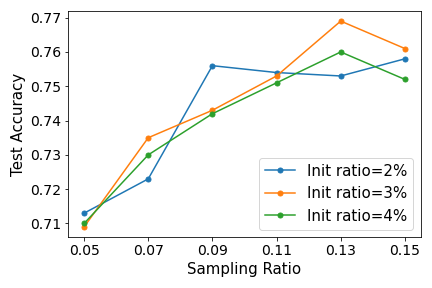}}
\caption{Compare different initial sampling ratios for Cora (left) and Citeseer (Right)}
\label{fig:init_ratio}
\end{figure}

\paragraph{Sensitivity to initial sampling ratio:}
During the initial sampling stage, \samplmethod samples $B_0$ nodes to train the model initially. And the initially trained model will affect the nodes sampled during the dynamic sampling period. We test the effect of different initial sampling ratio on Cora and Citeseer datasets. We vary the initial sampling ratio from $2\%$ to $4\%$, and Figure~\ref{fig:init_ratio} shows that \samplmethod is robust to the choice of initial sampling ratio. 

\paragraph{Diverse uncertainty algorithms:} Besides the sampling algorithm used by \samplmethod, there are some other algorithms to pick the representative nodes from the candidate set $S_t$. First, we will evaluate three algorithms to cluster and select the propagated features.
\begin{compactitem}
    \item Random select: randomly pick nodes $S_t$ from $\mathcal{C}_t$.
    \item \samplmethod: use $k$-means++ to cluster the nodes in $\mathcal{C}_t$ and 
    \item Random round-robin Algorithm~\cite{citovsky2021batch}: use the cluster labels from the initial sampling period (the initial sampling period clusters all the nodes in $\setV_{train}$). Then, following the Algorithm~\ref{alg:round-robin} to select $S_t$ from $\mathcal{C}_t$ 
\end{compactitem}

\begin{algorithm}[ht]
\caption{Random Round-robin Algorithm}
\label{alg:round-robin}
\begin{algorithmic}[1]
    \STATE Input: cluster labels of node $i$ (node $i \in \setV_{train}$) $cl_n$ , where $cl_n \in {1,2,\ldots,m}$; candidate set $\mathcal{C}_t$; number of nodes to label $B_t$.
    \STATE Using the cluster labels to split $\mathcal{C}_t$ onto clusters $A_1, A_2, \ldots, A_m$. Without loss of generality, $|A_1| \leq |A_2| \leq \ldots \leq |A_m|$. 
    \STATE $S_t = \emptyset$
    \FOR{$i =1,2,\ldots, B_t$}
    \FOR{$j=1,2,\ldots,m$}
    \IF{$A_j \neq \emptyset$}
    \STATE Uniformly select $x$ from $A_j$ at random
    \STATE $A_j \leftarrow A_j \setminus \{x\}$, $S_t \leftarrow S_t \cup \{x\}$
    \STATE break
    \ENDIF 
    \ENDFOR
    \ENDFOR
    \STATE return $S_t$
\end{algorithmic}
\end{algorithm}

\begin{figure}[H]
\centering
\subfloat[Cora]{\includegraphics[width=0.32\textwidth]{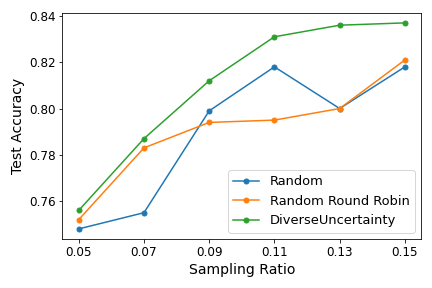}} 
\subfloat[Citeseer]{\includegraphics[width=0.32\textwidth]{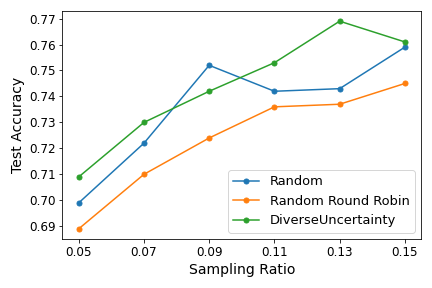}} 
\subfloat[Pubmed]{\includegraphics[width=0.32\textwidth]{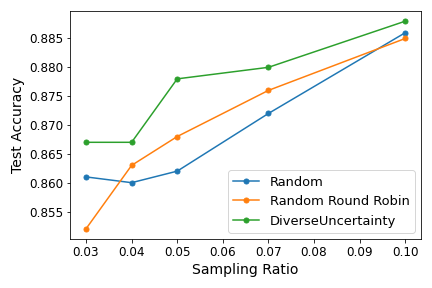}}
\caption{Compare different sampling algorithms to collect $S_t$ from the candidate set $\mathcal{C}_t$.}
\label{fig:comp_cluster_method}
\end{figure}

Figure~\ref{fig:comp_cluster_method} suggests that $k$-means++ clustering algorithm achieves a better test accuracy in most cases compared to random selection or random round-robin algorithm. Moreover, compared to random sampling algorithm, $k$-means++ clustering algorithm is more robust when the sampling ratio increases. As the sampling ratio increases, the test accuracy of $k$-means++ keeps increasing in most cases, while the test accuracy of random sampling algorithm has more fluctuations.

\begin{figure}[H]
\centering
\subfloat[Cora\label{fig:r=0.0}]{\includegraphics[width=0.32\textwidth]{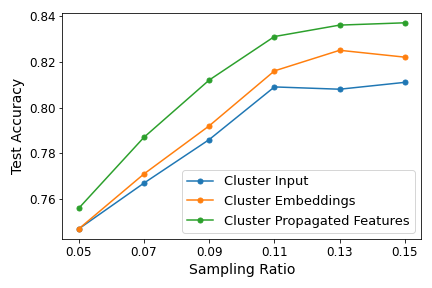}} 
\subfloat[Citeseer\label{fig:r=0.0}]{\includegraphics[width=0.32\textwidth]{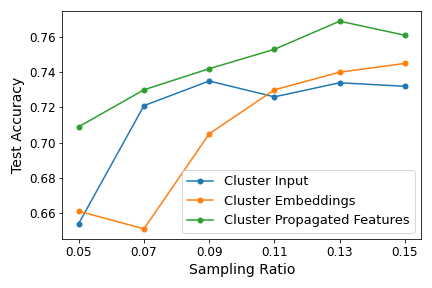}} 
\subfloat[Pubmed\label{fig:r=0.0}]{\includegraphics[width=0.32\textwidth]{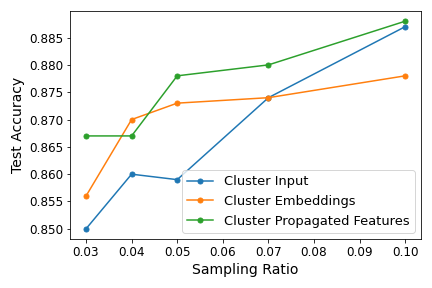}}
\caption{Compare clustering different targets to select $S_t$ from the candidate set $\mathcal{C}_t$.}
\label{fig:comp_cluster_target}
\end{figure}

Another factor that affects the test performance is the metric for clustering. Besides the propagated features (which is used by MaxDiversity), we can also cluster the input features or the embedding vectors. Since the GNN models typically used do not have a fully connected layer connecting to the output, we cannot use the output of second last layer as the embedding. Hence, we use the GNN output as the embedding vector for clustering. Figure~\ref{fig:comp_cluster_target} shows that clustering the propagated features consistently outperforms clustering the other two targets. Especially for the ``Citeseer'' dataset, clustering the propagated features outperforms by at most $5\%$. To conclude, the $k$-means++ clustering algorithm achieves the best performance compared to the other selection methods and clustering the propagated features is better than clustering other targets. Thus, \samplmethod uses $k$-means++ to cluster the propagated features to pick $S_t$ from $\mathcal{C}_t$.

\section{Empirical validation of theorem}

In this section, we perform simulation analysis to demonstrate that \method can reduce the MSE compared to greedy uncertainty sampling approach. 

\paragraph{Graph Simulation Setup}
Let the dimension of input feature $d=1$. Simulate $\mX$ from two different clusters, where $(X|C_1) \sim Uniform(-15, -5)$ and $(X|C_2) \sim Uniform(8, 12)$. In our simulation, we randomly generated 100 nodes for each cluster. Then, we simulate the edges between nodes. The edges can be divided into two categories, edges within clusters and edges between clusters. To simulate the edges within clusters, for each node, we random select two other nodes from the same cluster as its neighbor. For the edges between clusters, we set a probability threshold $r$ such that $\PR{V_i \in C_1 \text{ connect to a node} \in C_2}= r$. For each node $V_i \in C_1$, generate an indicator variable $I_i \sim Bernoulli(r)$ to determine whether $V_i$ is connected to cluster 2 ($V_i$ is connected to cluster 2 if $I_i=1$). If $V_i$ is connected to cluster 2, randomly pick a node from cluster 2 and connect it to $V_i$.

\paragraph{Label of nodes}
The label of a node depends on its propagated features. First compute the 1-layer feature propagation of each node, $\mX^{(1)}$. Then, the label of $i$-th node is $y_i = |X^{(1)}_i|^2$. Here, because the two cluster centers are equally distanced from 0, hence, the label function is also symmetric around $0$.

\paragraph{Node sampling}
During the initial sampling step, label the nodes closest to the cluster centers and train the GP function. To sample uncertain nodes,
\begin{compactitem}
    \item MaxUncertainty: Label the 8 nodes with largest uncertainty.
    \item \samplmethod: Collect the top 80 nodes with largest uncertainty into the candidate set. Then, use $k$-means++ to cluster the nodes in the candidate set into 8 clusters. Label the 8 nodes closest to the cluster centers. 
\end{compactitem}
MaxUncertainty and \samplmethod use the newly labeled nodes to update the GP function respectively. Finally,  the trained GP function predicts the node labels, and we compute the corresponding MSE.

\begin{figure}[ht]
\centering
\subfloat[MaxUncertainty ($r=0.0$)\label{fig:r=0.0}] {\includegraphics[width=0.49\textwidth]{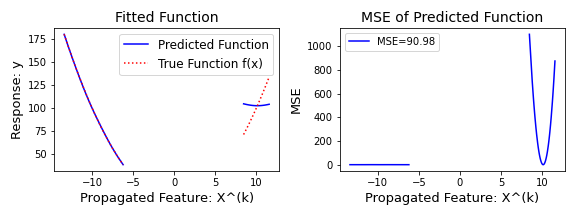}} \hfill
\subfloat[\samplmethod ($r=0.0$)\label{fig:r=0.0}] {\includegraphics[width=0.49\textwidth]{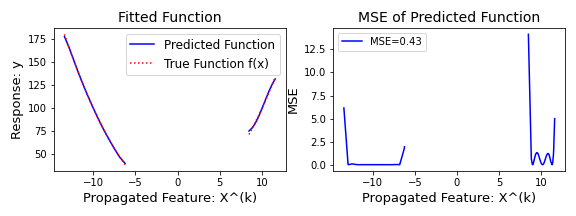}} \hfill
\subfloat[MaxUncertainty ($r=0.3$)\label{fig:r=0.8}]{\includegraphics[width=0.49\textwidth]{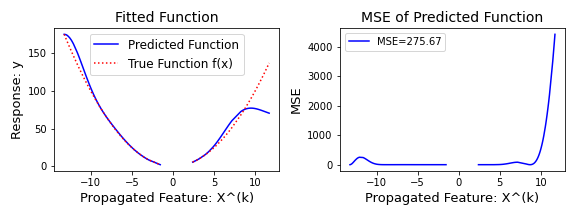}} \hfill
\subfloat[\samplmethod ($r=0.3$)\label{fig:r=0.8}] {\includegraphics[width=0.49\textwidth]{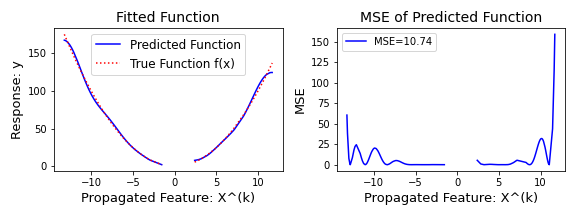}} \hfill
\subfloat[MaxUncertainty ($r=0.8$)\label{fig:r=0.8}] {\includegraphics[width=0.49\textwidth]{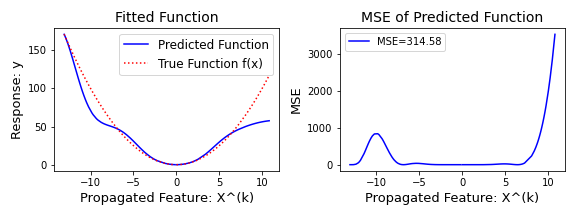}} \hfill
\subfloat[\samplmethod ($r=0.8$)\label{fig:r=0.8}] {\includegraphics[width=0.49\textwidth]{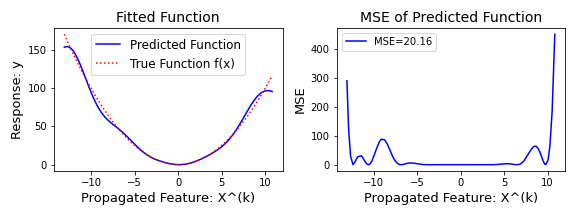}} \hfill
\caption{Compare the MSEs of Uncertainty and \samplmethod algorithms under different correlation levels between clusters. }
\label{fig:simu}
\end{figure}

Figure~\ref{fig:simu} suggests that MaxUncertainty has larger MSE compared to \samplmethod algorithm. For the MaxUncertainty algorithm, since most of the labeled nodes come from the cluster 1, the MSE of cluster 1 is significantly smaller than that of cluster 2. While for the \samplmethod algorithm, the MSE of cluster 1 and 2 are comparable. As $r$ increases, there are more and more edges between clusters, and the propagated features are less separated. Hence, there are some high uncertainty nodes from cluster 1 very close to cluster 2, which is beneficial for MaxUncertainty to learn the labels of nodes from cluster 2. Thus, we could observe $\frac{\text{MSE of MaxUncertainty}}{\text{MSE of \samplmethod}}$ keeps decreasing when $r$ increases. When $r$ is very large, cluster 1 and 2 will merge into one cluster, and MSEs of both methods no longer have a significant difference.

\section{Conclusion}

Learning a GNN model with limited labeling budget is an important but challenging problem. In this paper:
\begin{compactitem}
\item 
We propose a novel data efficient GNN learning algorithm, \method, which efficiently diversifies the uncertain nodes and achieves better test accuracy than recent  baselines. 
\item We  provide theoretical guarantees: Theorem~\ref{theorem_1}
 proves the advantage of \method over MaxUncertainty sampling. 
\item
Experiments on real data show that \method can save up to
50\% labeling size, for the same test accuracy.
\end{compactitem}
We envision \method will inspire future research of combining uncertainty sampling and representation sampling (diversifying).

\printbibliography

@inproceedings{wu2019simplifying,
  title={Simplifying graph convolutional networks},
  author={Wu, Felix and Souza, Amauri and Zhang, Tianyi and Fifty, Christopher and Yu, Tao and Weinberger, Kilian},
  booktitle={International conference on machine learning},
  pages={6861--6871},
  year={2019},
  organization={PMLR}
}

@article{kipf2016semi,
  title={Semi-supervised classification with graph convolutional networks},
  author={Kipf, Thomas N and Welling, Max},
  journal={arXiv preprint arXiv:1609.02907},
  year={2016}
}

@article{hu2020ogb,
  title={Open Graph Benchmark: Datasets for Machine Learning on Graphs},
  author={Hu, Weihua and Fey, Matthias and Zitnik, Marinka and Dong, Yuxiao and Ren, Hongyu and Liu, Bowen and Catasta, Michele and Leskovec, Jure},
  journal={arXiv preprint arXiv:2005.00687},
  year={2020}
}

@article{wu2019active,
  title={Active learning for graph neural networks via node feature propagation},
  author={Wu, Yuexin and Xu, Yichong and Singh, Aarti and Yang, Yiming and Dubrawski, Artur},
  journal={arXiv preprint arXiv:1910.07567},
  year={2019}
}

@article{zhang2021grain,
  title={Grain: Improving data efficiency of graph neural networks via diversified influence maximization},
  author={Zhang, Wentao and Yang, Zhi and Wang, Yexin and Shen, Yu and Li, Yang and Wang, Liang and Cui, Bin},
  journal={arXiv preprint arXiv:2108.00219},
  year={2021}
}

@inproceedings{defferrard2016convolutional,
  title={Convolutional neural networks on graphs with fast localized spectral filtering},
  author={Defferrard, Micha{\"e}l and Bresson, Xavier and Vandergheynst, Pierre},
  booktitle=NIPS,
  month={Dec.},
  address={Barcelona, Spain},
  pages={3844--3852},
  year={2016}
}

@article{bronstein2017geometric,
  title={Geometric deep learning: going beyond euclidean data},
  author={Bronstein, Michael M and Bruna, Joan and LeCun, Yann and Szlam, Arthur and Vandergheynst, Pierre},
  journal=IEEESPM,
  volume={34},
  number={4},
  pages={18--42},
  year={2017},
  publisher={IEEE}
}

@article{citovsky2021batch,
  title={Batch Active Learning at Scale},
  author={Citovsky, Gui and DeSalvo, Giulia and Gentile, Claudio and Karydas, Lazaros and Rajagopalan, Anand and Rostamizadeh, Afshin and Kumar, Sanjiv},
  journal={Advances in Neural Information Processing Systems},
  volume={34},
  year={2021}
}

@ARTICLE{ioanniditensorgcnstspjournal,
  author={V. N. {Ioannidis} and A. G. {Marques} and G. B. {Giannakis}},
  journal={IEEE Transactions on Signal Processing}, 
  title={Tensor Graph Convolutional Networks for Multi-Relational and Robust Learning}, 
  year={2020},
  volume={68},
  number={},
  pages={6535-6546},
  doi={10.1109/TSP.2020.3028495}}

@article{zhu2002learning,
  title={Learning from labeled and unlabeled data with label propagation},
  author={Zhu, Xiaojin and Ghahramani, Zoubin},
  year={2002},
  publisher={Citeseer}
}

@article{cai2017active,
  title={Active learning for graph embedding},
  author={Cai, Hongyun and Zheng, Vincent W and Chang, Kevin Chen-Chuan},
  journal={arXiv preprint arXiv:1705.05085},
  year={2017}
}

@inproceedings{gao2018active,
  title={Active discriminative network representation learning},
  author={Gao, Li and Yang, Hong and Zhou, Chuan and Wu, Jia and Pan, Shirui and Hu, Yue},
  booktitle={IJCAI International Joint Conference on Artificial Intelligence},
  year={2018}
}

@article{velivckovic2017graph,
  title={Graph attention networks},
  author={Veli{\v{c}}kovi{\'c}, Petar and Cucurull, Guillem and Casanova, Arantxa and Romero, Adriana and Lio, Pietro and Bengio, Yoshua},
  journal={arXiv preprint arXiv:1710.10903},
  year={2017}
}

@article{settles2009active,
  title={Active learning literature survey},
  author={Settles, Burr},
  year={2009},
  publisher={University of Wisconsin-Madison Department of Computer Sciences}
}

@article{hanneke2014theory,
  title={Theory of disagreement-based active learning},
  author={Hanneke, Steve and others},
  journal={Foundations and Trends{\textregistered} in Machine Learning},
  volume={7},
  number={2-3},
  pages={131--309},
  year={2014},
  publisher={Now Publishers, Inc.}
}

@article{tur2005combining,
  title={Combining active and semi-supervised learning for spoken language understanding},
  author={Tur, Gokhan and Hakkani-T{\"u}r, Dilek and Schapire, Robert E},
  journal={Speech Communication},
  volume={45},
  number={2},
  pages={171--186},
  year={2005},
  publisher={Elsevier}
}

@article{ducoffe2018adversarial,
  title={Adversarial active learning for deep networks: a margin based approach},
  author={Ducoffe, Melanie and Precioso, Frederic},
  journal={arXiv preprint arXiv:1802.09841},
  year={2018}
}

@techreport{arthur2006k,
  title={k-means++: The advantages of careful seeding},
  author={Arthur, David and Vassilvitskii, Sergei},
  year={2006},
  institution={Stanford}
}

@article{o1978curve,
  title={Curve fitting and optimal design for prediction},
  author={O'Hagan, Anthony},
  journal={Journal of the Royal Statistical Society: Series B (Methodological)},
  volume={40},
  number={1},
  pages={1--24},
  year={1978},
  publisher={Wiley Online Library}
}

@article{seeger2004gaussian,
  title={Gaussian processes for machine learning},
  author={Seeger, Matthias},
  journal={International journal of neural systems},
  volume={14},
  number={02},
  pages={69--106},
  year={2004},
  publisher={World Scientific}
}

@article{kirsch2019batchbald,
  title={Batchbald: Efficient and diverse batch acquisition for deep bayesian active learning},
  author={Kirsch, Andreas and Van Amersfoort, Joost and Gal, Yarin},
  journal={Advances in neural information processing systems},
  volume={32},
  pages={7026--7037},
  year={2019}
}

@article{wang2019dgl,
    title={Deep Graph Library: A Graph-Centric, Highly-Performant Package for Graph Neural Networks},
    author={Minjie Wang and Da Zheng and Zihao Ye and Quan Gan and Mufei Li and Xiang Song and Jinjing Zhou and Chao Ma and Lingfan Yu and Yu Gai and Tianjun Xiao and Tong He and George Karypis and Jinyang Li and Zheng Zhang},
    year={2019},
    journal={arXiv preprint arXiv:1909.01315}
}

@inproceedings{hamilton2017inductive,
  title={Inductive representation learning on large graphs},
  author={Hamilton, William L and Ying, Rex and Leskovec, Jure},
  booktitle={Proceedings of the 31st International Conference on Neural Information Processing Systems},
  pages={1025--1035},
  year={2017}
}

\end{document}